%% file: ACL2024/acl_latex.tex
\pdfoutput=1

\documentclass[11pt]{article}

\usepackage{acl}

\usepackage{times}
\usepackage{latexsym}

\usepackage[T1]{fontenc}

\usepackage[utf8]{inputenc}

\usepackage{microtype}

\usepackage{inconsolata}

\usepackage{graphicx}

\usepackage{physics}
\usepackage{mathtools,nccmath}
\usepackage{graphicx}
\usepackage{booktabs}
\usepackage{amsfonts,amssymb}
\usepackage{bm}
\usepackage{threeparttable}
\usepackage{multirow}
\usepackage{amsmath}
\usepackage{subfigure}
\usepackage{enumitem}
\usepackage{times}
\usepackage{soul}
\usepackage{url}
\usepackage{amsthm}
\usepackage{booktabs}
\usepackage{algorithm}
\usepackage{algorithmic}
\usepackage{caption}
\usepackage{CJKutf8}
\usepackage{hyperref}
\usepackage{subcaption}
\usepackage{arydshln}

\title{Self-Error-Instruct: Generalizing from Errors for \\ LLMs  Mathematical Reasoning}

\author{
  Erxin Yu\textsuperscript{1}, Jing Li\textsuperscript{1,2}\thanks{Corresponding author}, Ming Liao\textsuperscript{1}, Qi Zhu\textsuperscript{3},  
  Boyang Xue\textsuperscript{4}, Minghui Xu\textsuperscript{3}, \\ \textbf{Baojun Wang}\textsuperscript{3},  \textbf{Lanqing Hong}\textsuperscript{3},  
  \textbf{Fei Mi}\textsuperscript{3}, \textbf{Lifeng Shang}\textsuperscript{3} \\
  \textsuperscript{1}Department of Computing, The Hong Kong Polytechnic University \\
  \textsuperscript{2}Research Centre for Data Science \& Artificial Intelligence \\
  \textsuperscript{3}Huawei Noah’s Ark Lab, \textsuperscript{4}The Chinese University of Hong Kong \\
  \texttt{erxin.yu@connect.polyu.hk}, \texttt{jing-amelia.li@polyu.edu.hk}
}

\begin{document}
\maketitle
\input{ACL2024/tex_files/0abstract}
\input{ACL2024/tex_files/1introduction}
\input{ACL2024/tex_files/2relatedWork}

\input{ACL2024/tex_files/3method}
\input{ACL2024/tex_files/4experimentSetup}

\input{ACL2024/tex_files/5exprimentResult}

\input{ACL2024/tex_files/6conclusion}

\input{ACL2024/tex_files/7limitation}
\input{ACL2024/tex_files/ackownledgments}
\bibliography{anthology,custom}

\appendix
\input{ACL2024/tex_files/appendix}

\end{document}

%% file: ACL2024/tex_files/0abstract.tex
\begin{abstract}
    Although large language models demonstrate strong performance across various domains, they still struggle with numerous bad cases in mathematical reasoning.
    Previous approaches to learning from errors synthesize training data by solely extrapolating from isolated bad cases, thereby failing to generalize the extensive patterns inherent within these cases.
    This paper presents Self-Error-Instruct (SEI), a framework that addresses these model weaknesses and synthesizes more generalized targeted training data. 
    Specifically, we explore a target model on two mathematical datasets, GSM8K and MATH, to pinpoint bad cases. Then, we generate error keyphrases for these cases based on the instructor model's (GPT-4o) analysis and identify error types by clustering these keyphrases. 
    Next, we sample a few bad cases during each generation for each identified error type and input them into the instructor model, which synthesizes additional training data using a self-instruct approach.
    This new data is refined through a one-shot learning process to ensure that only the most effective examples are kept. Finally, we use these curated data to fine-tune the target model, iteratively repeating the process to enhance performance.
    We apply our framework to various models and observe improvements in their reasoning abilities across both in-domain and out-of-domain mathematics datasets.
    These results demonstrate the effectiveness of self-error instruction in improving LLMs' mathematical reasoning through error generalization. 

\end{abstract}

%% file: ACL2024/tex_files/1introduction.tex
\section{Introduction}

Large language models (LLMs) \citep{fewshotlearners, ChatGPT, mistral, gemini} have demonstrated remarkable capabilities across various domains, particularly after instruction-based fine-tuning. Yet, LLMs are still facing substantial challenges in complex reasoning tasks, particularly in mathematical reasoning. They continue to encounter numerous bad cases, often committing errors that compromise their reliability.

Previous work has taken advantage of these errors to improve model performance. 
Mistake-tuning and self-rethinking \citep{tong-etal-2024-llms} leverage the historical errors of LLMs to enhance their performance during both the fine-tuning and inference stages. LLMs like ChatGPT \citep{ChatGPT} are utilized to synthesize training datasets based on the bad cases from smaller models \citep{ying-etal-2024-llms,tong-etal-2024-securing}. LLMs are also employed to optimize the reasoning steps of smaller models \citep{LEMA}, generating corrective data to train these models.

However, current methods predominantly synthesize training data from individual bad cases. While this can somewhat enhance model performance, the data often suffers from a lack of generalization because it is too reliant on specific instances, which limits its ability to cover a wider array of error patterns. To overcome this limitation, we introduce the Self-Error-Instruct (SEI) framework, which aims to generalize training data based on error types instead of focusing solely on individual cases.
For example, in Figure \ref{figure:intro}, the left subfigure displays various error types of Qwen2.5-Math. We enhanced its mathematical reasoning by generalizing the data according to these error types, which is depicted in the right subfigure. To the best of our knowledge, \textit{we are the first to explore data synthesis and selection for LLMs to generalize from errors based on error types in math reasoning.}

\input{ACL2024/figures/intro}

Specifically, we begin by assessing target model to identify bad cases. An instructor model is first used to pinpoint errors from these bad cases and generate relevant keyphrases, then cluster these keyphrases into distinct error types. We select a few samples from each error type as prompts for the instructor model in a self-instruct manner to synthesize new data. We further apply a one-shot learning-based refinement to the new data to verify its effectiveness to rectify the target model's deficiencies while maintaining the target model's current success, only keeping the data that works. This refinement process is iteratively repeated to improve the model's performance.



We employ LLaMA3-8B-Instruct, Qwen2.5-Math-7B, and Mathstral-7B-v0.1 as the target models to identify bad cases within the training datasets, GSM8K and MATH. We conduct comprehensive evaluations using both in-domain and out-of-domain testing. For in-domain tests, we use test sets from GSM8K and MATH. For out-of-domain tests, we utilize four additional mathematical reasoning datasets: TAL, GaoKao, SAT, and College.

Experimental results show that training the target models with our synthesized data significantly improves performance on both in-domain and out-of-domain test sets. 
Specifically, LLaMA3 and Mathstral achieve average improvements of 1.72\% and 0.98\%, respectively, while Qwen2.5 shows a more significant gain of 24.94\%.
Additionally, our one-shot learning-based data selection method is highly effective, outperforming both random selection and LESS \citep{less}, a recently proposed gradient-based data selection method. 
It also surpasses the performance of models trained on the full dataset.
This demonstrates that our approach can accurately identify high-quality training data to enhance model performance. Our experiments further highlight the importance of resolving bad cases in the one-shot learning selection process and maintaining the model's correctness on the original good cases. 
Finally, we analyze the fix rate of bad cases at each iteration, examine the impact of generalized data volume on model performance, and compare two training strategies: iterative training with data synthesized in each round versus training from scratch with all synthesized data. In summary, our contributions are as follows:

$\bullet$ We improve data generalization by organizing mathematical reasoning data according to error types instead of individual bad cases.

$\bullet$ We propose the Self-Error-Instruct framework, which analyzes bad cases through keyphrases extraction and clustering, then performs data generalization for each cluster.

$\bullet$ Experiments show that our method efficiently generalizes data based on error types, enhancing mathematical reasoning skills and validating the effectiveness of our data selection strategy.

%% file: ACL2024/figures/intro.tex
\begin{figure*}
    \centering
    \subfigure{
        \begin{minipage}[t]{0.45\textwidth} 
        \centering
        \includegraphics[width=1\textwidth]{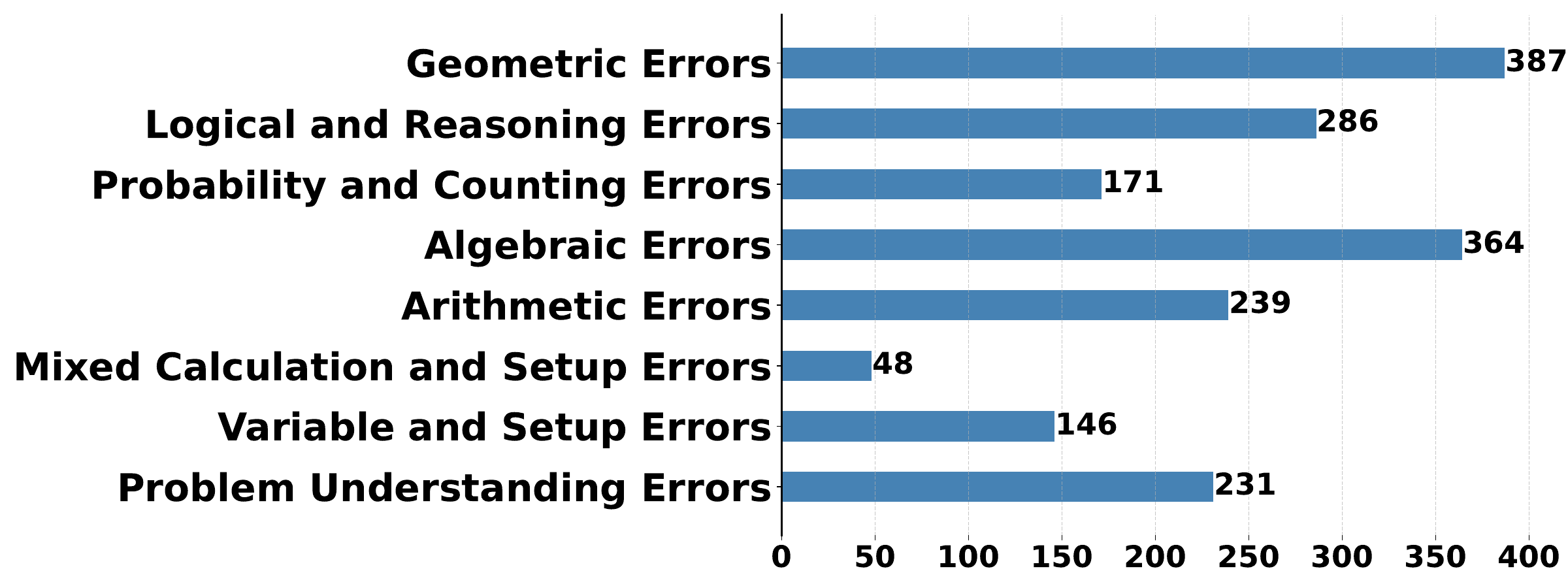}
        \end{minipage}%
    }\subfigure{
        \begin{minipage}[t]{0.45\textwidth} 
        \centering
        \includegraphics[width=1\textwidth]{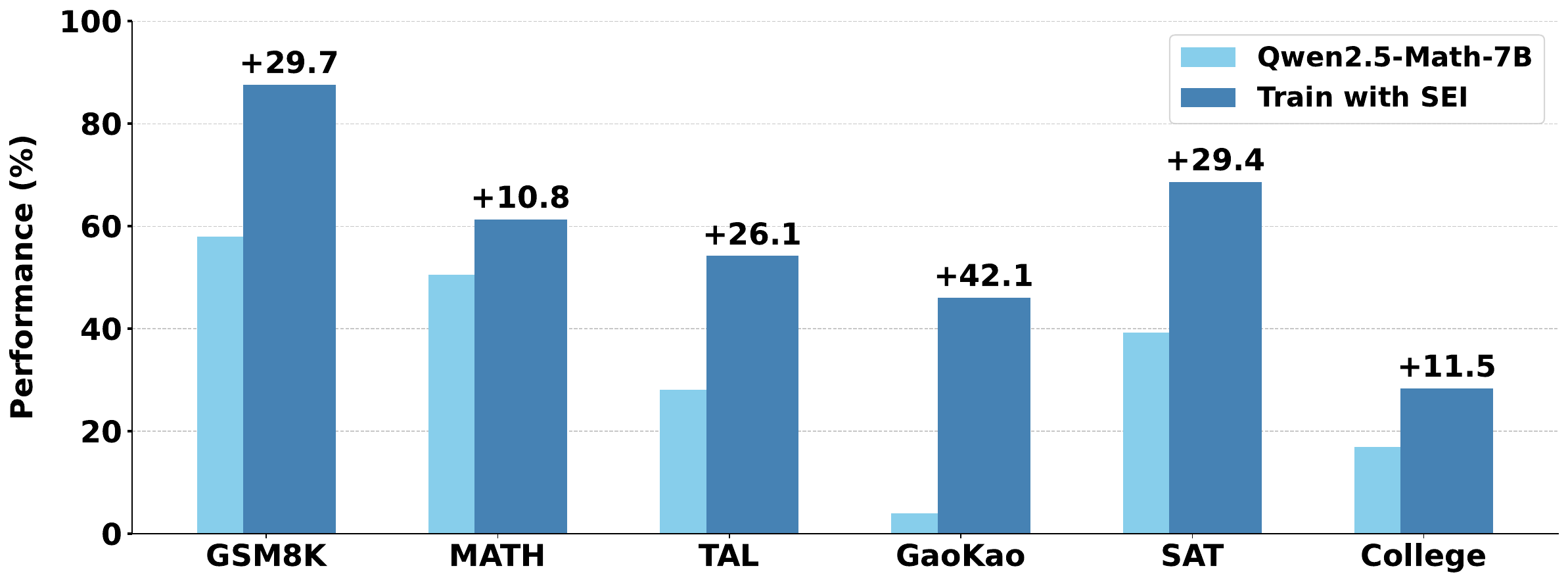}
        \end{minipage}%
    }
    \caption{The left table shows some error types of Qwen2.5-Math-7B on Math and GSM8K training set, while the right presents the results after training on data generalized from error categories.}
    \label{figure:intro}
    \vspace{-3mm}
\end{figure*}

%% file: ACL2024/tex_files/2relatedWork.tex
\section{Related Work}
\subsection{Mathematical Reasoning}
With the rapid advancement of large language models, they have shown remarkable capabilities across a wide range of NLP tasks, as demonstrated by models like ChatGPT \citep{ChatGPT}, Claude \citep{claude}, and Gemini \citep{gemini}. However, mathematical reasoning remains a significant challenge for these models. 
To address this issue, many models, such as OpenAI o1 \citep{o1}, Qwen-2.5-Math \citep{qwen2.5math}, and DeepSeek-Math \citep{deepseekmath}, have undergone specialized training for mathematical tasks.
Researchers have explored various strategies to enhance performance in this area, including prompting, pretraining, and fine-tuning.

Among these techniques, some focus specifically on learning from errors to enhance model performance.
LEMA \citep{LEMA} leveraged GPT-4 \citep{gpt4} to correct the model's erroneous reasoning paths and used the refined reasoning paths to fine-tune the model.
Self-rethinking and mistake tuning \citep{tong-etal-2024-llms} analyze the causes of model errors to improve reasoning performance. The former uses an iterative process to help the model avoid repeating past mistakes, while the latter fine-tunes the model by incorporating correct and erroneous reasoning examples.
LLM2LLM \citep{tong-etal-2024-securing} generates new synthetic data based on error cases to improve model performance iteratively. 
Learning from error and learning from error by contrast \citep{ying-etal-2024-llms} are two strategies designed to improve the performance of target models. The former generates targeted training data by analyzing erroneous responses, while the latter by contrasting correct and incorrect responses.
In contrast to these approaches, which focus solely on individual bad cases, our method generalizes data based on error types. This allows for more systematic coverage of diverse issues, enhances data diversity, and improves generalization ability.

\subsection{Data Selection}

Data selection plays a crucial role in instruction tuning, as it helps identify high-quality data, enhancing model performance and generalization while minimizing noise to optimize training.
LIMA \citep{lima} achieved exceptional performance by selecting 1,000 high-quality question-answer pairs for instruction tuning, delivering results comparable to those obtained through large-scale instruction tuning and reinforcement learning.
Instruction-following difficulty \citep{li-etal-2024-quantity} was proposed to evaluate the difficulty of following instructions for each sample. LESS \citep{less} identified training data most similar to the validation set based on gradient features. 
NUGGETS \citep{li-etal-2024-one} assessed the impact of candidate instructions on a predefined task set's perplexity using one-shot learning, comparing the score differences between zero-shot and one-shot learning as a reference for data selection.
Building on NUGGETS, we designed a one-shot learning data selection method tailored for mathematical reasoning. This method selects data based on whether the generated data can address the target model's bad cases while preserving its good cases.

%% file: ACL2024/tex_files/3method.tex
\section{Our Self-Error-Instruct Framework}
\input{ACL2024/figures/framework}
Our framework \footnote{Our code is available at \url{https://github.com/ErxinYu/SEI}.} aims to enhance the mathematical reasoning ability of the target model $\mathbf{M}_{\text{target}}$ by identifying its weaknesses, referred to as bad cases, on an existing mathematical training dataset $\mathbf{D}_{\text{train}}$. These bad cases are analyzed to guide the synthesis of targeted training data that directly addresses the model's specific shortcomings. By progressively training on this tailored data, the mathematical capabilities of $\mathbf{M}_{\text{target}}$ are effectively improved.
 
As shown in Figure \ref{table:main result}, our process consists of four key steps:  
1) \textbf{Bad Case Extraction} (\text{Section \ref{ssection:bad case extraction}}), which identifies the incorrect cases where the target model $\mathbf{M}_{\text{target}}$ fails on the existing mathematical reasoning dataset $\mathbf{D}_{\text{train}}$.  
2) \textbf{Self Error Instruct} (\text{Section \ref{ssection:Self Error Instruct}}) generates targeted data for $\mathbf{M}_{\text{target}}$ by first identifying error keyphrase, then clustering similar errors, and finally synthesizing data specifically tailored to address the identified error types.  
3) \textbf{Data Selection} (\text{Section \ref{ssection:data selection}}) filters and selects high-quality data from the generated dataset, ensuring that only the most relevant and effective examples are used for training.  
4) \textbf{Iterative Training} (\text{Section \ref{ssection:training}}) uses the selected data to retrain $\mathbf{M}_{\text{target}}$, iterating this process to continuously refine and enhance the model's performance, thereby improving its mathematical reasoning capabilities with each cycle.

\subsection{Bad Case Extraction}
\label{ssection:bad case extraction}
For each problem with its correct reasoning path $(q_i, r_i)$ in the training dataset $\mathbf{D}_{\text{train}}$, we use $\mathbf{M}_{\text{target}}$ to generate a reasoning path. During this process, we identify and collect the bad case $(q_i,  r_i, \hat{r_i})$ into the error dataset $\mathbf{D}_{\text{error}}$, where the answers derived from the reasoning paths differ, i.e., $\mathbf{Ans}(\hat{r_i}) \neq \mathbf{Ans}(r_i)$, where $\mathbf{Ans}(\cdot)$ is the function that extracts the answer from a given reasoning path. Thus, the error dataset is defined as:

\begin{equation}
\mathbf{D}_{\text {error}} = \{(q_i, r_i, \hat{r_i}) \mid \mathbf{Ans}(\hat{r_i}) \neq \mathbf{Ans}(r_i)\}.
\end{equation}

\subsection{Self Error Instruct}
\label{ssection:Self Error Instruct}
In this phase, for each bad case in $\mathbf{D}_{\text{error}}$, we leverage the $\mathbf{M}_{\text{instructor}}$ model to perform error analysis by examining the reasoning paths and generating an error keyphrase that captures the nature of the mistake. These error keyphrases are then clustered into distinct groups based on similarity. For each error type, targeted data synthesis generates new training samples specifically designed to address model weaknesses. This process produces the curated dataset $\mathbf{D}_{\text{SEI}}$, containing diversity and error-specific training samples to enhance the target model's reasoning ability.

\paragraph{Error Keyphrase Generation.}

During this stage, we address each bad case $(q_i, r_i, \hat{r_i})$ in the dataset $\mathbf{D}_{\text{error}}$ using the $\mathbf{M}_{\text{instructor}}$ model for detailed error analysis. This process generates an error keyphrase $e_i$, which captures the specific nature of the error. To achieve this, we employ a structured function $\mathbf{Extract}[\cdot]$ with a keyphrase extraction prompt to analyze the incorrect reasoning path $\hat{r_i}$ and produce the corresponding error keyphrase. Details of the prompt are provided in the Appendix \ref{appendix:keyphrase}. The process is mathematically represented as follows:

{\small
\begin{equation}
\begin{split}
\textit{EK-Set} = \big\{e_i \mid e_i = \mathbf{Extract}[\mathbf{M}_{\text{instructor}}, (q_i, r_i, \hat{r_i})], \\
\forall (q_i, r_i, \hat{r_i}) \in \mathbf{D}_{\text{error}} \big\},
\end{split}
\end{equation}
}where $\textit{EK-Set}$ represents the collection of error keyphrases generated for all bad cases in $\mathbf{D}_{\text{error}}$. This approach ensures that each $e_i$ accurately captures the underlying issue in the model's reasoning path, providing a solid foundation for subsequent clustering and data synthesis steps.

\paragraph{Error Keyphrases Clustering.}

After obtaining the \textit{EK-Set}, we utilize the $\mathbf{M}_{\text{instructor}}$ model to cluster the keyphrases within this set. This clustering process identifies distinct error types, denoted as the \textit{ET-Set}. The process can be mathematically expressed as:

\begin{equation}
\textit{ET-Set} = \mathbf{Cluster}[\mathbf{M}_{\text{instructor}}, \textit{EK-Set}],
\end{equation}
where $\mathbf{Cluster}[\cdot]$ is a clustering prompt (see Appendix \ref{appendix:generation}) designed to group the error keyphrases into coherent and distinct types. Each type is manually reviewed (see Appendix \ref{appendix:manul}) to filter and validate its relevance and appropriateness.

\paragraph{Error Type-Specific Data Synthesis.}

For each error type within the \textit{ET-Set}, we begin by sampling a subset of bad cases from the same error type, which serve as in-context learning prompts. These prompts are then used to guide $\mathbf{M}_{\text{instructor}}$ in generating additional data that falls under the same error type. This process ensures that the generated data remains consistent with the specific error patterns of the given type, thereby expanding our dataset with more diverse but relevant examples. Through this process, we ultimately obtain a synthesized dataset $\mathbf{D}_{\text{SEI}}$, which enriches our data with examples covering distinct error patterns. The specific prompt used for this generalization process can be found in the Appendix \ref{appendix:systhetic}.

\subsection{One-shot Learning Selection}
\label{ssection:data selection}
After obtaining the generalized dataset $\mathbf{D}_{\text{SEI}}$ targeting specific errors, our goal is to select a small subset of high-quality data for training the target model. In previous work,  NUGGETS \citep{li-etal-2024-one} uses a one-shot learning approach to filter data. It calculates a score for each instruction example based on its impact on the perplexity of a set of pre-defined tasks, allowing for the identification of the most beneficial data for instruction tuning. 

In our approach to mathematical reasoning tasks, instead of relying on perplexity, we directly evaluate whether the newly generalized data can effectively serve as a one-shot prompt to guide the target model in resolving bad cases. Furthermore, we aim to ensure that the target model maintains its performance on good cases originally answered correctly, preserving its effectiveness across challenging and straightforward examples. First, we randomly sample a subset of bad cases and good cases to create a validation set, $\mathbf{D}_{\text{dev}}$. 
Next, we evaluate each sample in $\mathbf{D}_{\text{SEI}}$ by measuring the number of cases in $\mathbf{D}_{\text{dev}}$ that can be resolved when the sample is used as a one-shot prompt.
This evaluation serves as the criterion for selecting high-quality data. The process can be represented as:

\begin{align}
r^{j}_i &= M_{\text{target}}(\underbrace{q^j r^j}_{\text{One-Shot Prompt}} \oplus q_i ) \\
S^{j}_{\text{osl}} &= \sum_{i} \mathbb{I}[Ans(r^{j}_i) = Ans(r_i)]
\end{align}

The expression $q^j r^j$ represents the $j$-th synthetic data point from the dataset $\mathbf{D}_{\text{SEI}}$. The score $S^{j}_{\text{osl}}$ is the one-shot learning score, calculated by summing the indicator function $\mathbb{I[\cdot]}$, which is 1 if the answer from $r^{j}_i$ matches $r_i$, and 0 otherwise. Here, $q_i r_i$ are elements from $\mathbf{D}_{\text{dev}}$, where $r_i$ is the correct reasoning path for $q_i$. The prompt for one-shot learning is shown in Appendix \ref{fig:one-shot}. For each synthetic data in $\mathbf{D}_{\text{SEI}}$, calculate the set of one-shot learning scores $\{S^{1}_{\text{osl}}, S^{2}_{\text{osl}}, \ldots, S^{m}_{\text{osl}}\}$. By sorting these scores, we obtain the selection $\mathbf{D}^{\text{osl}}_{\text{SEI}}$.

\subsection{Iterative Training Optimization}
\label{ssection:training}
The selected data, $\mathbf{D}^{\text{osl}}_{\text{SEI}}$, is used to train the target model, $\mathbf{M}_{\text{target}}$. After the model is enhanced through this training, it is applied to $\mathbf{D}_{\text{train}}$ once more to identify new bad cases that it still struggles with. This process is iterated, continuously optimizing the target model by improving its ability to handle challenging examples, thereby enhancing its overall mathematical reasoning ability.

%% file: ACL2024/figures/framework.tex
\begin{figure*}[htbp]
    \centering
    \includegraphics[width=0.95\textwidth]{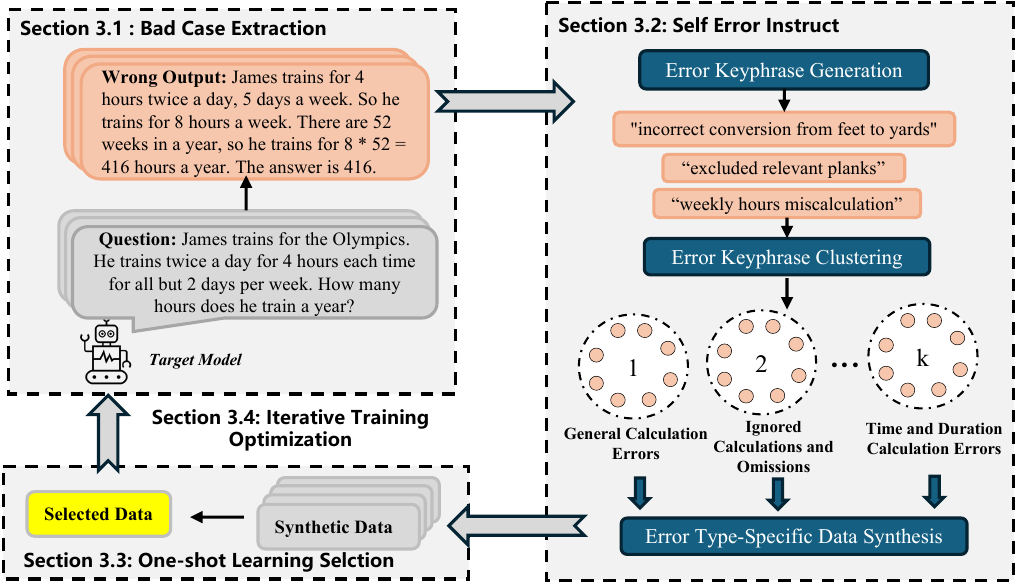} 
    \caption{An overview of our Self-Error-Instruct framework. It consists of four key steps: (1)\textbf{ Bad case extraction} identifies failure cases from the target model. (2) \textbf{Self-error-instruct} generates error keyphrases, clustering, and synthesizes data for each error type. (3) \textbf{One-shot learning data selection} retains only high-quality and effective examples for training. (4) \textbf{Iterative training} refines the target model by fine-tuning it with the curated data and repeating the process to further improve performance.} 
    \vspace{-3mm}
    \label{fig:framework} 
    \vspace{-3mm}
\end{figure*}

%% file: ACL2024/tex_files/4experimentSetup.tex
\section{Experimental Setup}
\subsection{Data Synthetic}
\input{ACL2024/tables/data_statistic}
\input{ACL2024/tables/main_result}

We identify bad cases from the training datasets of GSM8K and MATH, using GPT-4o \footnote{We use the Microsoft Azure AI services at \url{https://azure.microsoft.com/}} \cite{gpt4o} as the instructor model to generate error keyphrases, perform clustering, and synthesize data. For each error type, during the self-error instruct process, we sample 5 data points from the error dataset $\mathbf{D}_{\text{error}}$ and 3 data points from the already generated data within the current error type to serve as prompts. Each time, GPT-4o generalizes 20 new math data. We then filter out data with a Rouge-L score greater than 0.7 compared to the GSM8K and MATH training and test datasets to enhance diversity and prevent test set leakage. We randomly select 100 data points, comprising 50 good and 50 bad cases, to construct the validation set $\mathbf{D}_{\text{dev}}$. The number of iterations for data synthesis and model training is 3. 
In each iteration, we generate 10,000 data points by synthesizing 5,000 examples for the error types of GSM8K and 5,000 for MATH. We select the top 5\% of the synthetic data from each part and combine them into a unified dataset for training.
Over three iterations, we generate a total of 30,000 data points and select 1,500 for training. 
We also compared two methods for training the target model: iterative training, which starts from the model trained in the previous round, and training from scratch, which uses the selected data in a single step. The results of these two methods are shown in Table \ref{table:trainingMehotd}.

\subsection{Target Model Setting} 
We use the instruction-tuned Llama3-8b-instruct model \cite{llama3}, the math-specialized Qwen2.5-Math-7B \cite{qwen2.5math}, and Mathstral-7B-v0.1 \cite{mistral} as our target models. During training, we employ LoRA \citep{lora} with a maximum sequence length of 2048 tokens, set the number of training epochs to 3, and use a learning rate 2e-05.

\subsection{Evaluation} 
We used the GSM8K \citep{gsm8k} and Math \citep{MATH} test sets for in-domain evaluation. For out-of-domain evaluation, we utilized four challenging datasets: 1) \textbf{TAL-SCQ} \citep{TAL}: A K-12 mathematics test set containing 1,496 test examples. 2) \textbf{GaoKaoBench-Math} \citep{gaokao}: Comprising 508 test examples, this dataset features math problems from the Chinese high-school curriculum. 3) \textbf{SAT-MATH} \citep{SAT}: Consisting of 102 questions, this dataset includes math problems from the U.S. high-school curriculum. 4) \textbf{CollegeMath} \citep{college}: This dataset contains 2,818 test examples of college-level math problems. The detailed dataset statistics are provided in Table \ref{tab:data statistic}.

\input{ACL2024/tables/data_select}
We evaluated the models on these datasets using greedy decoding in a zero-shot setting, with the maximum generation length set to 2048. Performance was measured using Exact Match (EM), where answers were extracted from the generated reasoning paths and compared to the correct ones. All evaluations were conducted using the MWPBench framework \footnote{\url{https://github.com/microsoft/unilm/tree/master/mathscale/MWPBench}}.

\subsection{Baselines}

We compare with several baselines:  
1) \textbf{Training Data}, where the model is trained on the combined GSM8K and MATH datasets;  
2) \textbf{Bad Cases}, using bad cases from the initial target model;  
3) \textbf{LLMs-as-Instructors}, using Learning from error (LE) by generating tailored training data for errors. \citep{ying-etal-2024-llms} 
4) \textbf{Self-Instruct} \citep{wang-etal-2023-self-instruct}, generating 1,500 data points;  
7) \textbf{LLM2LLM} \citep{tong-etal-2024-securing}, also generating 1,500 data points;  
8) \textbf{Rand}, randomly selecting 500 data points per iteration for a total of 1,500; and  
9) \textbf{LESS} \citep{less}, selecting 1,500 data points based on gradient similarity.

We adopt the same setting as SEI for self-instruct, except that the sampled examples are selected randomly. Eight samples (five bad cases and three generated data) are selected in each iteration, and GPT-4o generates 20 new samples. This process is repeated to produce a total of 30,000 samples, from which 1,500 training samples are selected using the ICL method. For LLM2LLM and LLMs-as-Instructors, one new sample is generated per bad case using GPT-4o, with 500 samples generated per round over three rounds, resulting in 1,500 samples. We filter out samples with a Rouge-L similarity score above 0.7 during data synthesis by comparing them against the GSM8K and MATH training and test datasets.

For rand selection, data is proportionally sampled from each error type, with more samples drawn from types with more bad cases. For LESS, following the original setting, we randomly select 10 examples from GSM8K and MATH as the validation set, compute the average gradient of the validation set, and select generated data with the most similar gradients.

%% file: ACL2024/tables/data_statistic.tex
\begin{table}[htb!]
\centering
\resizebox{0.48\textwidth}{!}{%
\begin{tabular}{lcccc}
\toprule
\textbf{Dataset} & \textbf{Difficulty} & \textbf{Difficulty} & \textbf{Train} & \textbf{Test} \\ 
\midrule
GSM8K            & Elementary           & Easy           & 7,473 & 1,319           \\ 
MATH             & Competition          & ExHard         & 7,498 & 5,000           \\ 
TAL-SCQ          & K12 Math             & Medium         & -    & 1,496           \\ 
GaoKaoBech-Math  & High School          & Hard           & -    & 508            \\ 
SAT-MATH         & High School          & Hard           & -    & 102            \\ 
CollegeMath      & College              & ExHard         & -    & 2,818           \\ 
\bottomrule
\end{tabular}%
}
\caption{Statistics of Different Datasets. We extract bad cases from the GSM8K and MATH training sets and use the test sets of all datasets for evaluation. Datasets marked with ``-'' indicate only test data is available and are used for out-of-domain evaluation.}
\label{tab:data statistic}
\vspace{-4mm}
\end{table}

%% file: ACL2024/tables/main_result.tex
\begin{table*}[!ht]
    \centering
    \resizebox{0.9\textwidth}{!}{%
    \begin{tabular}{lccccccc}
        \toprule
        \multirow{2}*{\bf Models} & \multicolumn{2}{c}{\textbf{In-Domain}} & \multicolumn{4}{c}{\textbf{Out-of-Domain}}&\multirow{2}*{\bf AVG} \\
        \cmidrule(r){2-3} \cmidrule(r){4-7} 
        & \textbf{GSM8K} & \textbf{MATH} & \textbf{TAL} & \textbf{GaoKao} & \textbf{SAT} & \textbf{College} \\
        
        
        \midrule
        \textit{\textbf{Llama3-8B-Instruct}} &77.56 &27.36 &37.03 &15.55 &39.22 &15.54 &35.38\\
        \textbf{+ Training data} &63.99 &23.32 &29.01 &12.00 &34.31 &13.41 &29.34\\
        \textbf{+ Bad Cases}&65.13 &23.20 &30.08 &11.22 &33.33 &13.41 &29.40\\  
        \textbf{+ Self-Instruct}       &74.83 &26.20 &35.44 &14.76 &37.25 &15.26 &33.96\\  
        \textbf{+ LLMs-as-Instructors} &79.37 &27.84 &36.17 &16.14 &38.24 &\textbf{15.79 }&35.59\\
        \textbf{+ LLM2LLM}             &76.61 &27.60 &\textbf{40.10} &15.16 &38.24 &15.51 &35.54\\
        \textbf{+ SEI-ICL} &\textbf{79.76}&\textbf{28.42}&39.91&\textbf{16.73}&\textbf{42.15}&15.61&\textbf{37.10}\\
        
        \midrule
        \textbf{\textit{Qwen2.5-Math-7B}} &57.92&50.52&28.07&3.93&39.22&16.96&32.77\\
        \textbf{+ Training data} &57.54&56.22&46.19&38.78&65.69&24.20&48.10\\
        \textbf{+ Bad Cases} &64.21&56.90&45.45&34.44&63.73&22.36&47.85\\
        \textbf{+ Self-Instruct} &80.57&58.24&52.66&43.31&65.69&26.87&54.56\\
        \textbf{+ LLMs-as-Instructors} &79.31&58.76&\textbf{54.62}&45.43&63.73&28.07&54.99\\
        \textbf{+ LLM2LLM} &81.17&58.88&53.56&43.11&65.69&27.96&55.06\\
        \textbf{+ SEI-ICL} &\textbf{87.64}&\textbf{61.28}&54.21&\textbf{46.06}&\textbf{68.62}&\textbf{28.42}&\textbf{57.71}\\
        
        \midrule
        \textbf{\textit{Mathstral-7B-v0.1}} &80.67&52.58&48.66&47.83&61.76&\textbf{25.80}&52.88\\
        \textbf{+ Training data}            &72.10&44.40&41.44&42.91&56.86&24.17&46.98\\
        \textbf{+ Bad Cases}                &70.58&46.06&41.24&43.11&59.80&24.59&47.56\\
        \textbf{+ Self-Instruct}            &79.68&52.02&47.13&44.69&58.82&25.28&51.27\\
        \textbf{+ LLMs-as-Instructors}      &79.61&52.42&48.13&43.31&\textbf{63.73}&25.19&52.07\\
        \textbf{+ LLM2LLM}                  &81.35&52.64&46.79&45.87&59.08&25.16&51.82\\
         \textbf{+ SEI-ICL}&\textbf{82.87}&\textbf{53.70}&\textbf{49.47}&\textbf{48.62}&62.75&25.72&\textbf{53.86}\\


        \bottomrule
    \end{tabular}
    }
    \caption{Main results on in-domain and out-of-domain mathematical test sets, evaluated using the exact match (EM). All experiments are conducted in a zero-shot setting. SEI-ICL refers to our proposed method, which leverages the self-error-instruct framework to generalize and train using the top 5\% of data selected through one-shot learning. For fair comparison, the generalized data sizes for the baselines are kept consistent with SEI-ICL.}
    \label{table:main result}
    \vspace{-5mm}
\end{table*}

%% file: ACL2024/tables/data_select.tex
\begin{table*}[!ht]
    \centering
    \resizebox{\textwidth}{!}{%
    \begin{tabular}{lcccccccc}
        \toprule
        \multirow{2}*{\bf Models}&\multirow{2}*{\bf \# Samples} & \multicolumn{2}{c}{\textbf{In-Domain}} & \multicolumn{4}{c}{\textbf{Out-of-Domain}}&\multirow{2}*{\bf AVG} \\
        \cmidrule(r){3-4} \cmidrule(r){5-8} 
        &   & \textbf{GSM8K} & \textbf{MATH} & \textbf{TAL} & \textbf{GaoKao} & \textbf{SAT} & \textbf{College} \\
        \midrule
        \textbf{Llama-3-8B-Instruct} &-&77.56 &27.36 &37.03 &15.55 &39.22 &15.54 &35.38\\
        \textbf{SEI-FULL} &100\%             &78.01 &28.02 &38.64 &15.94 &41.18 &\textbf{16.25} &36.34\\
        \textbf{~ -Rand }   & 5\% (1,500)    &77.80 &\textbf{28.54} &37.43 &15.16 &40.20 &15.72 &35.81\\  
        \textbf{~ -LESS}  &5\% (1,500)       &77.95 &28.18 &36.83 &14.96 &39.22 &15.87 &35.50\\
        \hdashline
        \multirow{3}*{\bf ~ -One-shot ICL}  &5\% (1,500)&79.76&28.42&\textbf{39.91}&\textbf{16.73}&\textbf{42.15}&15.61&\textbf{37.10} \\ 
         \textbf{}&10\% (3,000)               &\textbf{79.98 }&27.96 &39.37 &15.75 &40.19 &16.22 &36.58\\
        \textbf{}&20\% (6,000)               &79.37 &28.18 &39.65 &15.94 &39.22 &15.51 &36.31\\

        \midrule
        \textbf{\textit{Qwen2.5-Math-7B}} &-&57.92&50.52&28.07&3.93&39.22&16.96&32.77\\
        \textbf{SEI-FULL}&100\%       &83.45&60.34&53.57&44.61&67.65&28.22&56.30\\   
        \textbf{~ -Rand} &5\% (1,500) &82.52&58.82&53.44&43.58&65.69&27.81&55.31\\
        \textbf{~ -LESS} &5\% (1,500) &83.13&59.76&53.69&45.28&66.67&28.14&56.11\\
        \hdashline
        \multirow{3}*{\bf ~ -One-shot ICL} &5\% (1,500) &\textbf{87.64}&61.28&54.21&\textbf{46.06}&\textbf{68.62}&28.42&\textbf{57.71}\\
        \textbf{}&10\%(3,000)         &85.74&\textbf{61.56}&\textbf{54.89}&45.76&65.69&28.33&57.16\\
        \textbf{}&20\% (6,000)        &86.58&60.78&54.76&44.29&63.73&\textbf{28.57}&56.45\\
        \midrule
        
        \textbf{\textit{Mathstral-7B-v0.1}} &-&80.67&52.58&48.66&47.83&61.76&\textbf{25.80}&52.88\\
        \textbf{SEI-FULL} &100\% &81.12&53.56&49.13&\textbf{49.61}&59.80&25.62&53.14\\   
        \textbf{~ -Rand} &5\% (1,500) &79.98&52.50&48.21&47.05&60.78&25.19&52.29\\
        \textbf{~ -LESS} &5\% (1,500) &79.68&52.20&48.60&48.03&60.78&25.23&52.42\\
        \hdashline
        \multirow{3}*{\bf ~ -One-shot ICL} &5\% (1,500) &82.87&\textbf{53.70}&49.47&48.62&62.75&25.72&\textbf{53.86}\\
        \textbf{} &10\% (3,000) &80.52&53.50&48.79&48.23&61.76&24.88&52.95\\
        \textbf{} &20\% (6,000) &\textbf{83.24}&53.40&\textbf{49.53}&46.85&\textbf{63.73}&24.77&53.59\\

        \bottomrule
    \end{tabular}
    }
    \caption{Model performance under different data selection strategies and samples. The bolded results highlight the best performance achieved using the FULL dataset and the top 5\% of samples selected through Rand, LESS, and one-shot ICL methods.}
    \label{table:data select}
    \vspace{-5mm}
\end{table*}

%% file: ACL2024/tex_files/5exprimentResult.tex
\section{Experimental Results}
\subsection{Main Results}

Table \ref{table:main result} presents our main results, from which we can draw several conclusions. 1) Our method, SEI-ICL, outperforms others by substantial margins in all math datasets. 
Specifically, after training, Llama-3-8B-Instruct improves by 1.72\% and Mathstral by 0.98\%, while Qwen2.5-Math-7B achieves an improvement of 24.94\%, highlighting the effectiveness of our error-type-guided data generation approach.
2) Training solely on the original GSM8K and MATH datasets or the identified bad cases results in performance degradation for the Llama3 and Mathstral models. This suggests that existing math training datasets offer limited benefits for already instruction-tuned models.
It highlights the necessity of data synthesis. 3) With the same amount of data, our data generation method outperforms other baselines. As shown in Table \ref{table:main result}, the average improvement achieved by SEI-ICL on all the models is higher than that of these baselines. Furthermore, combined with the results in Table \ref{table:data select}, we observe that even without data selection, randomly selecting the same amount of data (Rand) performs better than self-instruct (random generation), LLMs-as-Instructors and LLM2LLM (based on a single bad case), demonstrating that our error-type-guided data generation is more effective.

\subsection{Data Selection}

Table \ref{table:data select} presents the results of different data selection methods. By selecting the top 5\% of the data using our one-shot learning method, the performance of the trained models on target models surpasses that of SEI-FULL, which uses the full dataset for training. Furthermore, our models continue to outperform SEI-FULL as the amount of selected data increases. Under the same data size, the one-shot learning method achieves better results than rand selection and LESS, shows the effectiveness of the one-shot learning approach specifically designed for mathematical problem selection.
\input{ACL2024/figures/2zhuzhuang}
\input{ACL2024/figures/234_zhexian_gsm_math}
We conducted analysis experiments on the data selection validation set $\mathbf{D}_{\text{dev}}$ mentioned in Section \ref{ssection:data selection}. Specifically, we compared the approach of using only bad cases as $\mathbf{D}_{\text{dev}}$ with the combined approach that includes both good and bad cases. The results of these experiments are shown in Figure \ref{fig:ablation}. It can be observed that the combined approach outperforms the method using only bad cases across most datasets. This demonstrates that, when performing one-shot learning for data selection, it is important to ensure that the generated data addresses bad cases effectively and to maintain the correctness of the original good cases.

\subsection{Iterative Improvement Result}
\input{ACL2024/tables/iterion}
Table \ref{table:iteration} presents the bad case fix rate and test set performance of the Qwen2.5-Math model across different iterations. As shown, with the increase in iterations, the bad case fix rate consistently improves for both datasets, accompanied by a steady improvement in test set performance. This indicates that our method effectively identifies the model's error types in each iteration and generates targeted data for training, thereby enhancing the model's overall performance.

\subsection{Iterative vs. From-scratch Training}
\input{ACL2024/tables/trainingMethod}

Table \ref{table:trainingMehotd} highlights the differences between iterative training and from-scratch training within our framework. In iterative training, each new iteration continues training the target model obtained in the previous round. In contrast, from-scratch training involves directly training the initial target model once the data is obtained after three rounds of data generation.
\
The results show that from-scratch training outperforms iterative training. A possible explanation for this is that in each round of iterative training, we only select the top 5\% of the data for training. With such a small amount of data, iterative fine-tuning may lead to overfitting over multiple rounds. On the other hand, training from scratch aggregated datasets helps mitigate this issue, resulting in better overall performance.

\subsection{Different Synthetic Size}

We conducted an analysis between the amount of unfiltered synthetic data and performance, with the results presented in Figure \ref{figure:data size}. It can be observed that for all target models, the size of the generalization data is not proportional to performance. 
For Llama3, performance initially improves but eventually starts to decline. Specifically, the best performance on GSM8K is achieved with 15,000 training samples, while the optimal result on MATH is reached with 25,000 samples. In contrast, the results for Qwen2.5 and Mathstral are relatively inconsistent.
These findings further highlight the importance of data selection. For models like Llama3 and Mathstral, which have already undergone extensive instruction tuning, the quantity of data may not be the key to improving performance. Instead, the focus should shift to constructing small but high-quality datasets.

%% file: ACL2024/figures/2zhuzhuang.tex
\begin{figure}[tb!]
    \centering
    \includegraphics[width=0.5\textwidth]{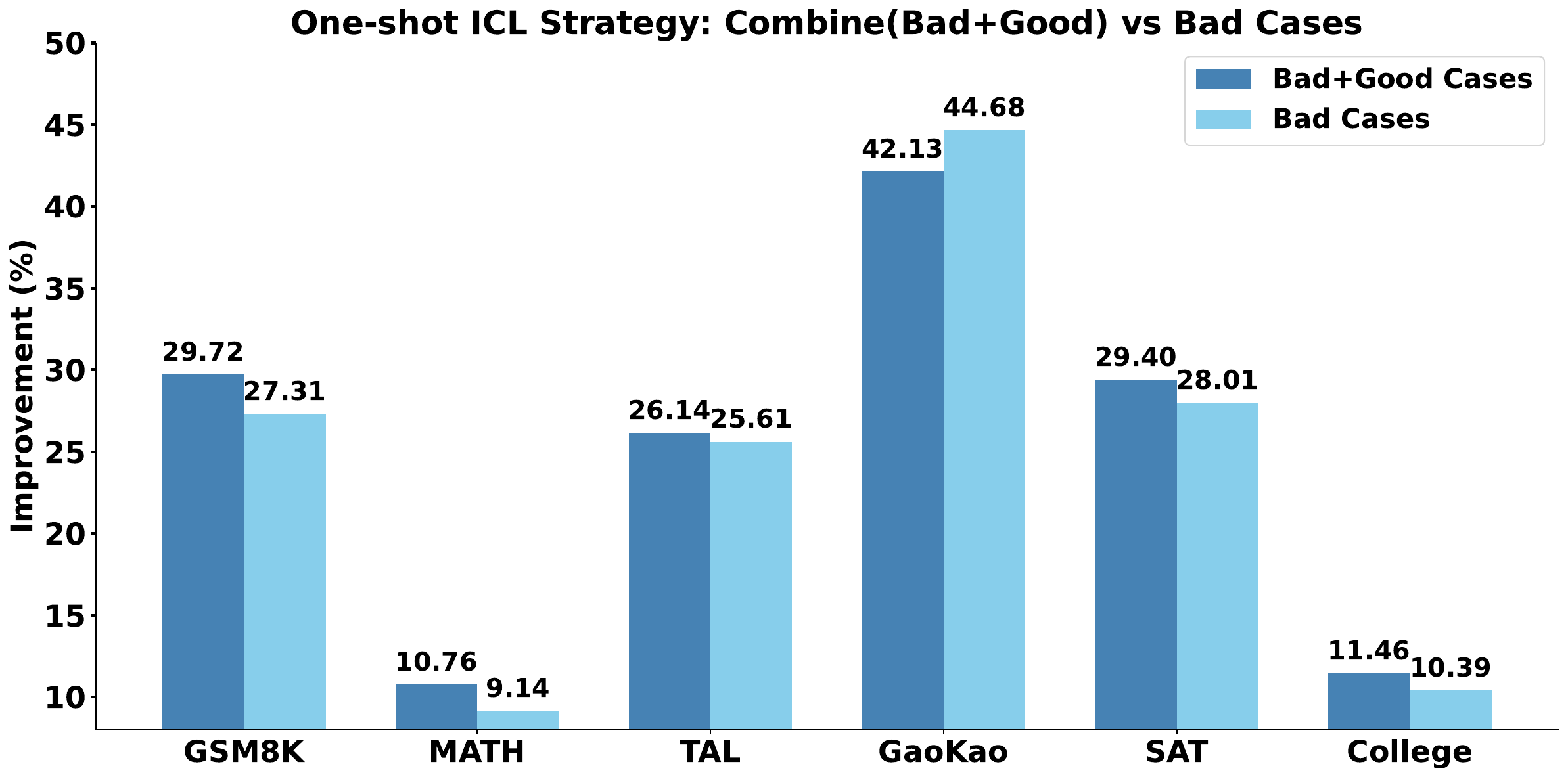} 
    \caption{The effects of two one-shot ICL strategies on the improvement of Qwen2.5.} 
    \label{fig:ablation} 
    \vspace{-3mm}
\end{figure}

%% file: ACL2024/figures/234_zhexian_gsm_math.tex
\begin{figure*}
    \centering
    \subfigure{
        \begin{minipage}[t]{0.45\textwidth} 
        \centering
        \includegraphics[width=1\textwidth]{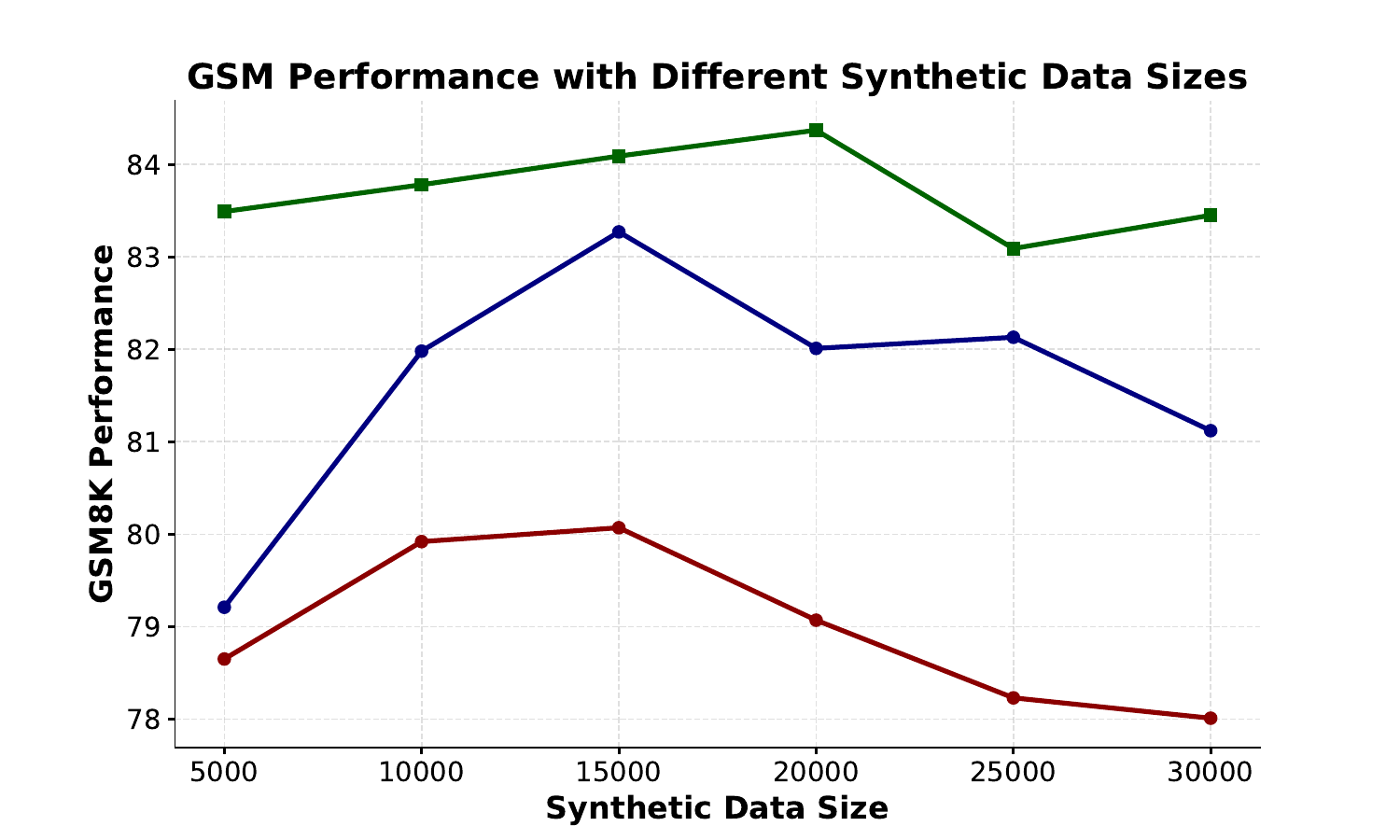}
        \label{figure:gsm8k_performance}
        \vspace{-6mm}
        \end{minipage}%
    }\subfigure{
        \begin{minipage}[t]{0.45\textwidth} 
        \centering
        \includegraphics[width=1\textwidth]{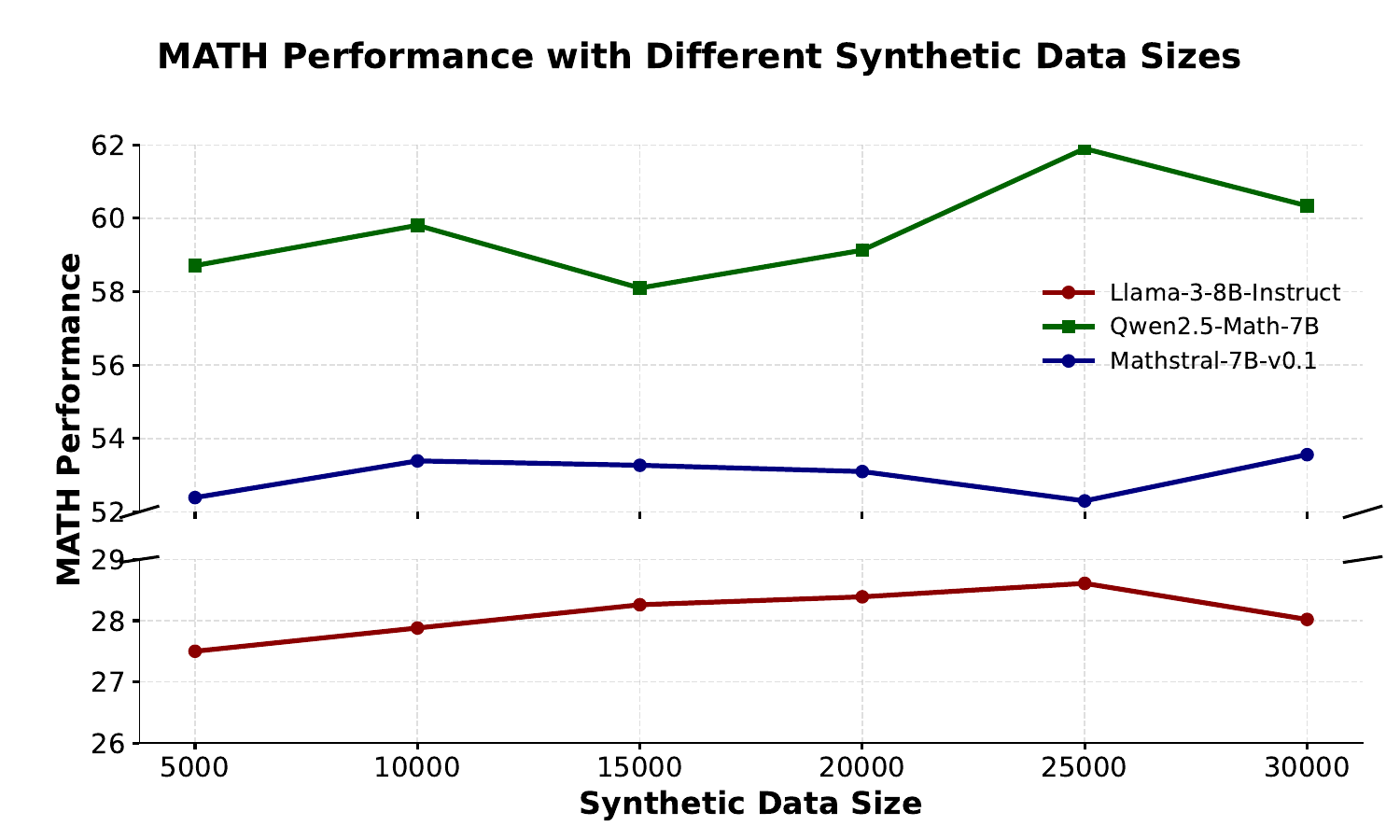}
        \label{figure:math_performance}
        \vspace{-6mm}
        \end{minipage}%
    }
    \caption{Comparison of GSM8K and MATH performance under different synthetic data sizes.}
    \label{figure:data size}
    \vspace{-3mm}
\end{figure*}


%% file: ACL2024/tables/iterion.tex
\begin{table}[!ht]
    \centering
    \resizebox{0.48\textwidth}{!}{%
    \begin{tabular}{lcccc}
        \toprule
        & \multicolumn{2}{c}{\textbf{Bad Case (Fix Rate)}} & \multicolumn{2}{c}{\textbf{Testset (EM Score)}} \\
        \cmidrule(r){2-3} \cmidrule(r){4-5} 
         & \textbf{GSM8K} & \textbf{MATH} & \textbf{GSM8K} & \textbf{MATH} \\
        \midrule
        \textbf{Iter-0 (ori)} &0 &0 &55.50&32.32 \\
        \textbf{Iter-1} &29.98 &23.17 &79.48 &57.21 \\
        \textbf{Iter-2} &38.01 &39.44&84.70 &58.19\\
        \textbf{Iter-3} &\textbf{39.13} &\textbf{40.57} &\textbf{87.79}&\textbf{59.18}\\

        \bottomrule
    \end{tabular}
    }
    \caption{Bad Case Fix Rate of Qwen2.5-Math on GSM8K and MATH during iterative improvement, along with its performance on the test sets. Bad cases refer to the errors made by Qwen2.5-Math in the training data of GSM8K and MATH.}
    \label{table:iteration}
\end{table}

%% file: ACL2024/tables/trainingMethod.tex
\begin{table}[!ht]
    \centering
    \resizebox{0.49\textwidth}{!}{%
    \begin{tabular}{lcccc}
        \toprule
        & \multicolumn{2}{c}{\textbf{GSM8K}} & \multicolumn{2}{c}{\textbf{MATH}} \\
        \cmidrule(r){2-3} \cmidrule(r){4-5}
        \textbf{Model} & \textbf{Iterative} & \textbf{From-scratch} & \textbf{Iterative} & \textbf{From-scratch} \\
        \midrule
        \textbf{Llama3} & 78.09 & \textbf{79.76 }& 27.62 & \textbf{28.42} \\
        \textbf{Qwen2.5} & \textbf{87.79} & 87.64 & 59.18 & \textbf{61.28 }\\
        \textbf{Mathstral} & 81.96 & \textbf{82.87} & 48.02 & \textbf{53.70} \\
        \bottomrule
    \end{tabular}
    }
    \caption{Comparison of model performance on GSM8K and MATH tasks under different training methods (Iterative and From-scratch).}
    \label{table:trainingMehotd}
\end{table}

%% file: ACL2024/tex_files/6conclusion.tex
\section{Conclusion}

We propose Self-Error-Instruct, a novel framework to improve LLMs mathematical reasoning by generalizing training data based on error types rather than individual bad cases. Our method enhances data diversity and mitigates overfitting by analyzing errors, clustering them into categories, and synthesizing targeted data using a self-instruct approach. 
Experiments on LLaMA3-8B-Instruct, Qwen2.5-Math-7B, and Mathstral demonstrate notable performance improvements with our method, achieving average gains of 1.72\%, 24.94\%, and 0.98\%, respectively, across both in-domain and out-of-domain evaluations.

%% file: ACL2024/tex_files/7limitation.tex
\section*{Limitations}
Our framework has three main limitations: the high cost of using GPT-4o as the instructor model, the focus on GSM8K and MATH datasets for bad case extraction, which may limit the diversity of errors, and the increased time consumption caused by one-shot learning.

Our approach is the reliance on GPT-4o as the instructor model for error analysis and data synthesis. While GPT-4o is highly effective in identifying error keyphrases and generating targeted training data, its use incurs significant computational and financial costs, which may limit the scalability and accessibility of the framework. 

The second limitation lies in the scope of our bad case extraction and iterative refinement process, which is currently confined to the GSM8K and MATH datasets. As a result, the error types identified and addressed may be limited to those specific to these datasets, potentially restricting the generalizability of the framework to other mathematical reasoning tasks or datasets. In the future, a more dynamic approach could be adopted, where bad cases are extracted from the initial datasets and continuously identified within the synthesized data during the iterative process. This would allow the framework to discover new and diverse error types as the training data evolves, further broadening the issues addressed and enhancing the model's mathematical reasoning capabilities. This expansion would help ensure the framework adapts to various problems, improving its robustness and applicability to real-world scenarios.

The third limitation lies in the one-shot data selection process. Although this approach is a one-time operation and produces results superior to LESS and random selection, the one-shot learning phase requires significant computational resources. This is because each of the 30,000 generated samples needs to be validated against an ICL-formatted validation set containing 100 samples.

\section*{Ethics Considerations}
This study strictly uses OpenAI's GPT-4o model for research purposes, in compliance with OpenAI's Business Terms, Section 2-(e). Our work analyzes reasoning errors to improve AI models and does not involve developing or commercializing competing products. We ensure no derived models are distributed or made available to third parties, maintaining full adherence to ethical and legal standards.


%% file: ACL2024/tex_files/ackownledgments.tex
\section*{Acknowledgements}
This work is supported by a grant from the Research Grants Council of the Hong Kong Special Administrative Region, China (Project No. PolyU/25200821), the Innovation and Technology Fund (Project No. PRP/047/22FX), PolyU Internal Fund from RC-DSAI (Project No. 1-CE1E), and a gift fund from Huawei (N-ZGM3).

%% file: ACL2024/tex_files/appendix.tex
\section{Overview of Prompts Used}
\subsection{Prompt for Training and Inference}
\label{appendix:alpaca}
For all the models, we use the built-in chat templates for training and inference. 
Figure \ref{fig:one-shot} illustrates the one-shot learning prompt for the Qwen2.5 model, where the model generates a response by being presented with an example of a synthetic question paired with its solution.

\subsection{Prompt for Error Keyphrase Generation}
\label{appendix:keyphrase}

Figure \ref{fig:keyphrase generation} illustrates the prompt used to generate error keyphrases for identifying and summarizing mistakes in mathematical reasoning. The input to the prompt includes a math question, the correct reasoning path leading to the answer, and the model's incorrect reasoning path. The prompt instructs the model to analyze where the error occurred in its reasoning process, identify the cause, and summarize it as a concise yet descriptive keyphrase. The output is a single keyphrase in list format, effectively capturing the primary reason for the model's mistake, which can then be used for further error analysis and targeted data synthesis.

\input{ACL2024/figures/keyphrasePropmt}
\subsection{Prompt for Error Clustering Generation}
\label{appendix:generation}
\input{ACL2024/figures/clusteringPrompt}
Figure \ref{fig:keyphrase cluster} presents a prompt designed to guide the analysis and categorization of error keyphrases generated from a model's reasoning mistakes. The input to this prompt is a list of error keyphrases, and the task involves clustering these keyphrases based on common themes, causes, or areas of occurrence. For each cluster, the model is instructed to list the included keyphrases, explain their grouping, and assign a concise, descriptive name to the cluster. This process helps identify patterns in the model's errors, offering meaningful insights into the types of mistakes made and enabling targeted improvements in the model's reasoning capabilities.

\subsection{Prompt for Error Type-Specific Data Synthesis}
\label{appendix:systhetic}
\input{ACL2024/figures/one-shot}
\input{ACL2024/figures/gsmSynPrompt}
\input{ACL2024/figures/mathSyntheticPrompt}
The prompt in Figure \ref{fig:gsm gen} and \ref{fig:math gen} guides the creation of 20 challenging math problems targeting specific error types in the GSM8K and MATH datasets. By analyzing the examples provided, the instruct model identifies patterns or issues causing errors and generates diverse, difficult problems aligned with these error types. The output follows a strict JSON format with detailed solutions and final numerical answers.

\section{Related Work on Data Synthesis}

The generation of synthetic data driven by large language models has become an essential method for addressing the issues of data quantity and quality in the field of deep learning \cite{survey}. 
LLMs, with their powerful language understanding and generation capabilities, can produce synthetic data that closely resembles the characteristics and patterns of real-world data \cite{wang-etal-2023-self-instruct}. 
This synthetic data can not only serve as a substitute or supplement for real data but can also be generated according to specific instructions and conditions to meet the needs of different applications \cite{repalm}. The use of LLM-driven synthetic data generation is widespread across various fields, including general alignment \cite{chen2024wapiti,cosafe,hallucination1}, mathematical reasoning \cite{lee-etal-2024-llm2llm,ying-etal-2024-llms}, medical \cite{midical1,xu-etal-2024-knowledge}, social media \cite{wei2025instructrag, popalm}, and hallucination \cite{hallucination3,hallucination2}.

\section{Manual Category Review}
\label{appendix:manul}
We applied two manual adjustments after clustering: \textbf{merging categories} and \textbf{excluding categories}. 

During the clustering process, some duplicate or similar categories may be generated, such as  ``Timezone and Duration Calculation Errors'' and ``Time and Duration Calculation Errors,'' or ``Calculation Errors'' and ``General Calculation Errors.'' These categories essentially represent the same or closely related error types, so we merge them into a unified category to avoid redundancy. 

We identify bad cases by comparing the model's extracted answers with the correct ones. However, this method may lead to a small number of correct answers being mistakenly identified as errors, which is a common issue in math evaluations. Fortunately, GPT-4o is usually able to determine that these responses are actually correct. Consequently, a special category like ``No Error'' or ``Correct Process'' may appear after clustering, and we manually exclude this category because it does not represent actual error types. Through these manual reviews, we can more accurately organize and analyze error categories, ensuring the reliability and consistency of the results. 

%% file: ACL2024/figures/keyphrasePropmt.tex
\begin{figure*}[htbp]
    \centering
    \includegraphics[width=1\textwidth]{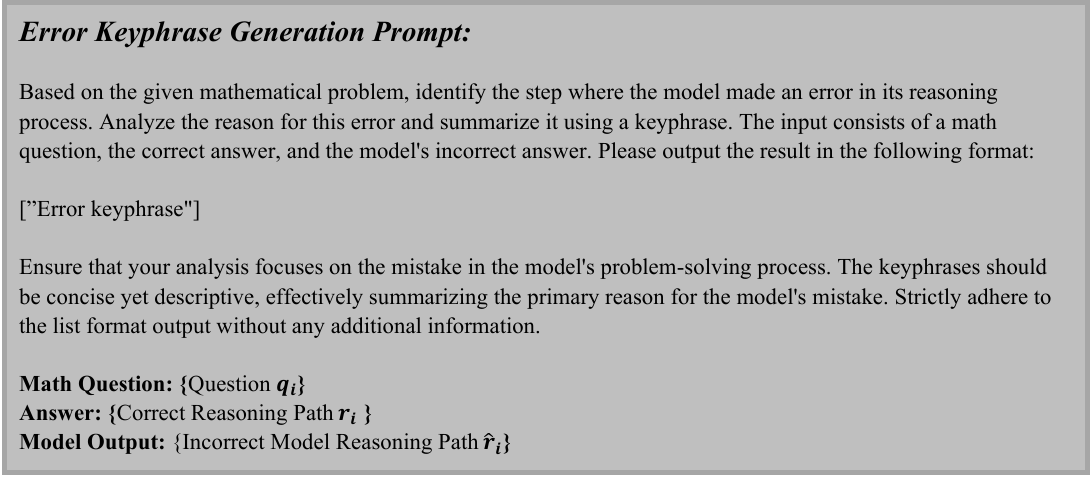} 
    \caption{Prompt for Generating Error Keyphrases.} 
    \label{fig:keyphrase generation} 
    \vspace{-5mm}
\end{figure*}

%% file: ACL2024/figures/clusteringPrompt.tex
\begin{figure*}[htbp]
    \centering
    \includegraphics[width=1\textwidth]{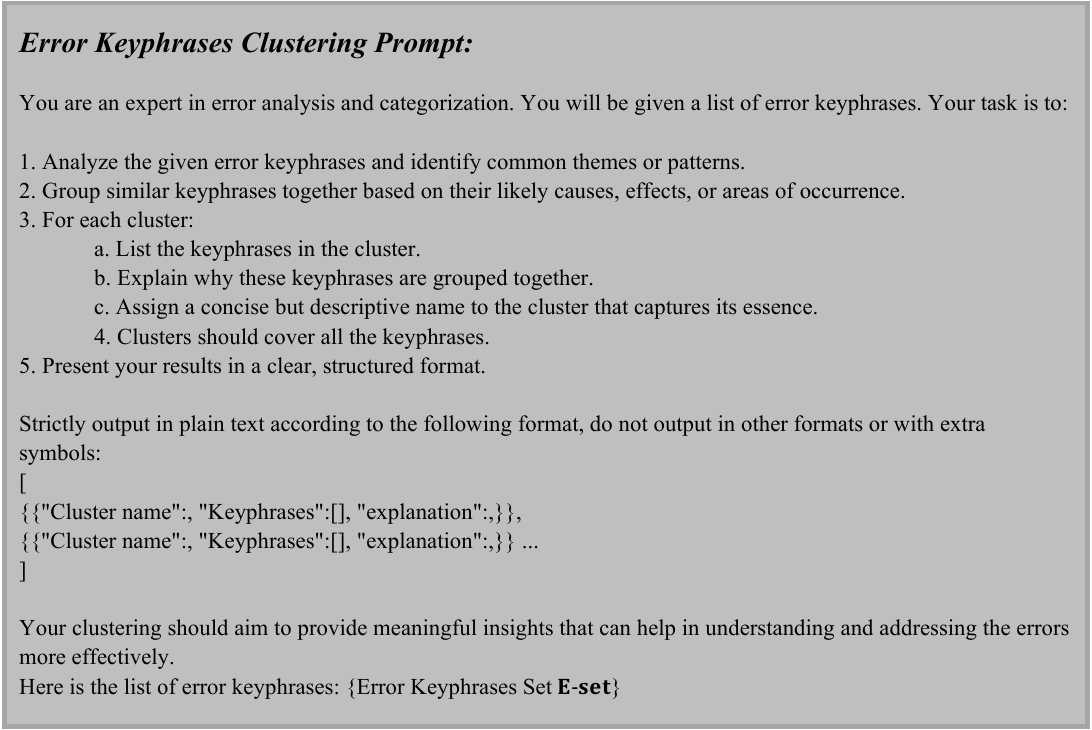} 
    \caption{Prompt for Clustering Error Keyphrases} 
    \label{fig:keyphrase cluster} 
\end{figure*}

%% file: ACL2024/figures/one-shot.tex
\begin{figure}[htbp]
    \centering
    \includegraphics[width=0.5\textwidth]{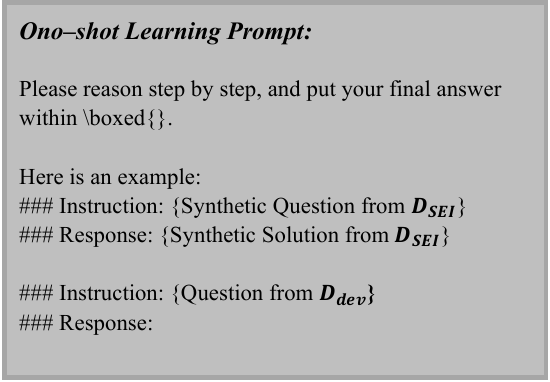} 
    \caption{One-Shot Learning Prompt for Selecting Synthetic Data} 
    \label{fig:one-shot} 
\end{figure}

%% file: ACL2024/figures/gsmSynPrompt.tex
\begin{figure}[htbp]
    \centering
    \includegraphics[width=0.45\textwidth]{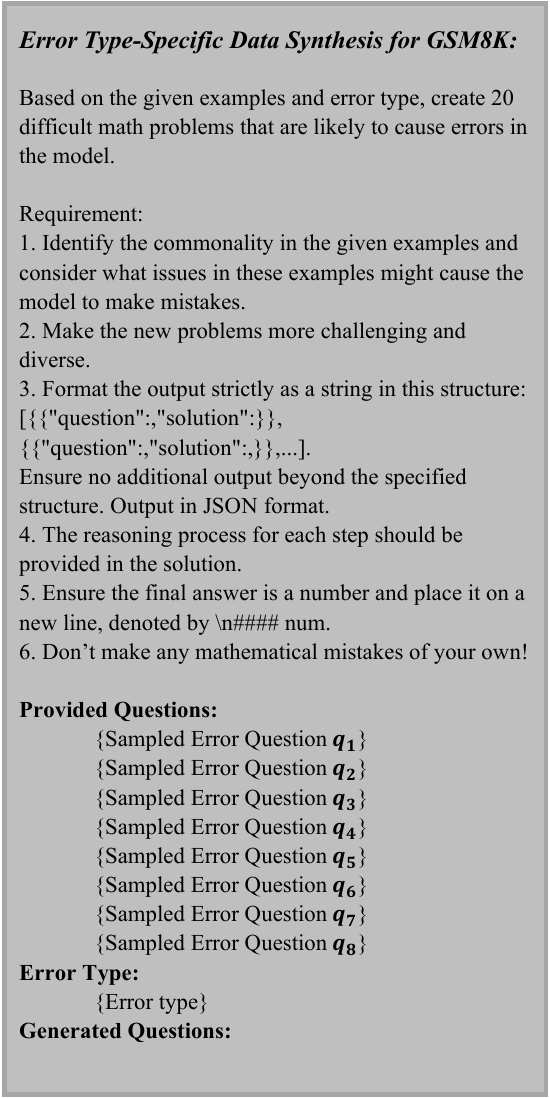} 
    \caption{Prompt for GSM8K Error Type-Specific Data Synthetic.} 
    \label{fig:gsm gen} 
\end{figure}

%% file: ACL2024/figures/mathSyntheticPrompt.tex
\begin{figure}[htbp]
    \centering
    \includegraphics[width=0.45\textwidth]{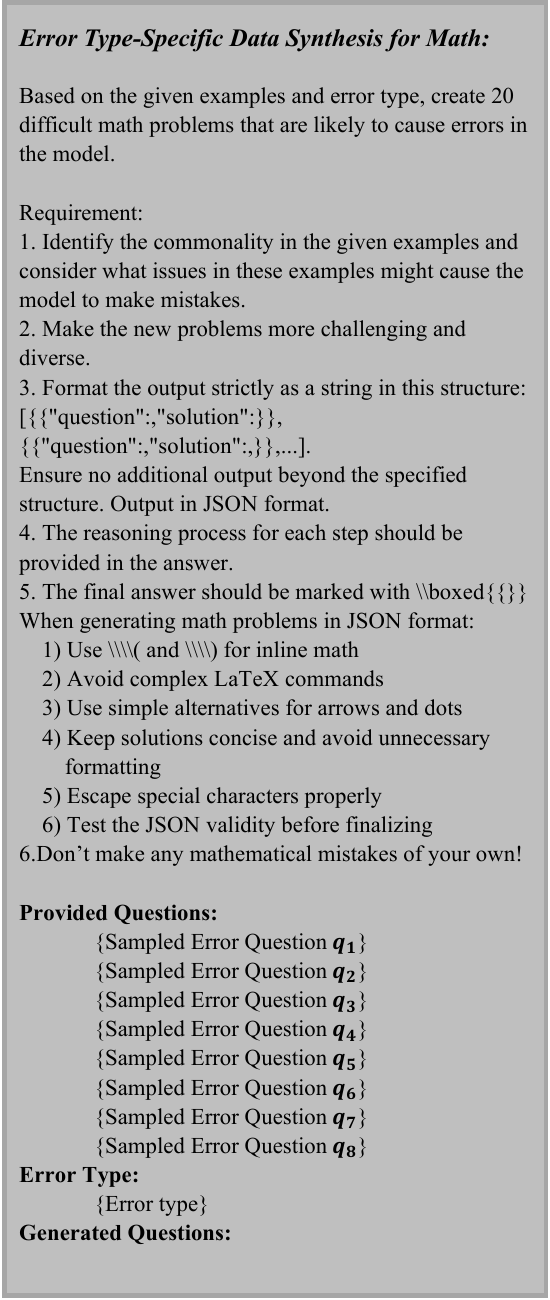} 
    \caption{Prompt for MATH Error Type-Specific Data Synthetic.} 
    \label{fig:math gen} 
\end{figure}